\documentclass{article}

\usepackage{algorithmic}
\usepackage{algorithm}
\usepackage{color}
\usepackage{amsmath}
\usepackage{amssymb}
\usepackage{fancyhdr}
\usepackage{graphicx}
\usepackage{makecell}
\usepackage{natbib}
\usepackage{subfigure} 
\usepackage{times}

\usepackage{swivel}

\begin{document} 
\twocolumn[
\mytitle{Swivel: Improving Embeddings by Noticing What's Missing}

\myauthor{Noam Shazeer}{noam@google.com}
\myauthor{Ryan Doherty}{portalfire@google.com}
\myauthor{Colin Evans}{colinhevans@google.com}
\myauthor{Chris Waterson}{waterson@google.com}
\myaddress{Google Inc., 1600 Amphitheatre Parkway, Mountain View, CA 94043}
\vskip 0.3in
]

\begin{abstract} 
We present \emph{Submatrix-wise Vector Embedding Learner} (Swivel), a
method for generating low-dimensional feature embeddings from a
feature co-occurrence matrix. Swivel performs approximate
factorization of the point-wise mutual information matrix via
stochastic gradient descent. It uses a piecewise loss with special
handling for unobserved co-occurrences, and thus makes use of
\emph{all} the information in the matrix. While this requires
computation proportional to the size of the entire matrix, we make use
of vectorized multiplication to process thousands of rows and columns
at once to compute millions of predicted values. Furthermore, we
partition the matrix into shards in order to parallelize the
computation across many nodes. This approach results in more accurate
embeddings than can be achieved with methods that consider only
observed co-occurrences, and can scale to much larger corpora than can
be handled with sampling methods.
\end{abstract} 

\section{Introduction}
\label{introduction}

Dense vector representations of words have proven to be useful for
natural language tasks such as determining semantic similarity,
parsing, and translation. Recently, work by \citet{word2vec} and
others has inspired an investigation into the construction of word
vectors using stochastic gradient descent methods. Models tend to fall
into one of two categories: matrix factorization or sampling from a
sliding window: \citet{dontcount} refers to these as ``count'' and
``predict'' methods, respectively.

In this paper, we present the \emph{Submatrix-wise Vector Embedding
  Learner} (Swivel), a ``count-based'' method for generating
low-dimensional feature embeddings from a co-occurrence matrix. Swivel
uses stochastic gradient descent to perform a weighted approximate
matrix factorization, ultimately arriving at embeddings that
reconstruct the point-wise mutual information (PMI) between each row
and column feature. Swivel uses a piecewise loss function to
differentiate between observed and unobserved co-occurrences.

Swivel is designed to work in a distributed environment. The original
co-occurrence matrix (which may contain millions of rows and millions
of columns) is ``sharded'' into smaller submatrices, each containing
thousands of rows and columns. These can be distributed across
multiple workers, each of which uses vectorized matrix multiplication
to rapidly produce predictions for millions of individual PMI
values. This allows the computation to be distributed across a cluster
of computers, resulting in an efficient way to learn embeddings.

This paper is organized as follows. First, we describe related word
embedding work and note how two popular methods are similar to one
another in their optimization objective.  We then discuss Swivel in
detail, and describe experimental results on several standard word
embedding evaluation tasks. We conclude with analysis of our results
and discussion of the algorithm with regard to the other approaches.

\section{Related Work}

While there are a number of interesting approaches to creating word
embeddings, Skipgram Negative Sampling \citep{word2vec} and GloVe
\citep{glove} are two relatively recent approaches that have received
quite a bit of attention.  These methods compress the distributional
structure of the raw language co-occurrence statistics, yielding
compact representations that retain the properties of the original
space. The intrinsic quality of the embeddings can be evaluated in two
ways. First, words with similar distributional contexts should be near
to one another in the embedding space. Second, manipulating the
distributional context directly by adding or removing words ought to
lead to similar translations in the embedded space, allowing
``analogical'' traversal of the vector space.

\textbf{Skipgram Negative Sampling.} The \texttt{word2vec} program
released by \citet{word2vec} generates word embeddings by sliding a
window over a large corpus of text. The ``focus'' word in the center
of the window is trained to predict each ``context'' word that
surrounds it by 1) maximizing the dot product between the sampled
words' embeddings, and 2) minimizing the dot product between the focus
word and a randomly sampled non-context word. This method of training
is called \emph{skipgram negative sampling} (SGNS).

\citet{matrixfactorization} examine SGNS and suggest that the
algorithm is implicitly performing weighted low-rank factorization of
a matrix whose cell values are related to the \emph{point-wise mutual
  information} between the focus and context words. Point-wise mutual
information (PMI) is a measure of association between two events,
defined as follows:

\begin{equation}
\mathbf{pmi}(i;j) = \log \frac{P(i,j)}{P(i)\,P(j)}
\end{equation}

In the case of language, the frequency statistics of co-occurring
words in a corpus can be used to estimate the probabilities that
comprise PMI. Let $x_{ij}$ be the number of times that focus word $i$
co-occurs with the context word $j$, $x_{i*} = \sum_j x_{ij}$ be total
number of times that focus word $i$ appears in the corpus, $x_{*j} =
\sum_i x_{ij}$ be the total number of times that context word $j$
appears appears in the corpus, and $\lvert D \rvert = \sum_{i,j}
x_{ij}$ be the total number of co-occurrences.  Then we can re-write
(1) as:

\begin{align*}
  \mathbf{pmi}(i;j)
  &= \log \frac{x_{ij} \lvert D \rvert}{x_{i*} \, x_{*j}} \\
  &= \log x_{ij} + \log \vert D \rvert - \log x_{i*} - \log x_{*j}
\end{align*}

It is important to note that, in the case that $x_{ij}$ is zero --
i.e., no co-occurrence of $i$ and $j$ has been observed -- PMI is
infinitely negative.

SGNS can be seen as producing two matrices, $\mathbf{W}$ for focus
words and $\mathbf{\tilde{W}}$ for context words, such that their
product $\mathbf{W} \mathbf{\tilde{W}}^\top$ approximates the observed
PMI between respective word/context pairs. Given a specific focus word
$i$ and context word $j$, SGNS minimizes the magnitude of the
difference between $w_i^\top \tilde{w}_j$ and $\mathbf{pmi}(i; j)$,
tempered by a monotonically increasing weighting function of the
observed co-occurrence count, $f(x_{ij})$:

\begin{align*}
\mathcal{L}_{\mathrm{SGNS}}
&= \sum_{i,j} f(x_{ij}) \left( w_i^\top \tilde{w}_j - \mathbf{pmi}(i;j) \right)^2 \\
&= \sum_{i,j} f(x_{ij}) ( w_i^\top \tilde{w}_j - \log x_{ij} - \log \lvert D \rvert \\
& \;\;\;\;\;\;\;\;\;\;\;\;\;\;\;\;\;\;\;\;\;\;\;\;+ \log x_{i*} + \log x_{*j} )^2
\end{align*}

Due to the fact that SGNS slides a sampling window through the entire
training corpus, a significant drawback of the algorithm is that it
requires training time proportional to the size of the corpus.

\textbf{GloVe.}  \citeauthor{glove}'s \citeyear{glove} GloVe is an
approach that instead works from the precomputed corpus co-occurrence
statistics. The authors posit several constraints that should lead to
preserving the ``linear directions of meaning''.  Based on ratios of
conditional probabilities of words in context, they suggest that a
natural model for learning such linear structure should minimize the
following cost function for a given focus word $i$ and context word
$j$:

\begin{align*}
  \mathcal{L}_{\mathrm{GloVe}} &= \sum_{i,j} f(x_{ij}) \left( w_i^\top \tilde{w}_j - \log x_{ij} + b_i + b_j \right)^2
\end{align*}

Here, $b_i$ and $b_j$ are bias terms that are specific to each focus
word and each context word, respectively.  Again $f(x_{ij})$ is a
function that weights the cost according to the frequency of the
co-occurrence count $x_{ij}$. Using stochastic gradient descent, GloVe
learns the model parameters for $\mathbf{W}$, $\mathbf{b}$,
$\mathbf{\tilde{W}}$, and $\mathbf{\tilde{b}}$: it selects a pair of
words observed to co-occur in the corpus, retrieves the corresponding
embedding parameters, computes the loss, and back-propagates the error
to update the parameters. GloVe therefore requires training time
proportional to the number of observed co-occurrence pairs, allowing
it to scale independently of corpus size.

Although GloVe was developed independently from SGNS (and, as far as
we know, without knowledge of \citeauthor{matrixfactorization}'s
\citeyear{matrixfactorization} analysis), it is interesting how
similar these two models are.

\begin{itemize}

\item Both seek to minimize the difference between the model's
  estimate and the log of the co-occurrence count. GloVe has
  additional free ``bias'' parameters that, in SGNS, are pegged to the
  corpus frequency of the individual words.  Empirically, it can be
  observed that the bias terms are highly correlated to the frequency
  of the row and column features in a trained GloVe model.

\item Both weight the loss according to the frequency of the
  co-occurrence count such that frequent co-occurrences incur greater
  penalty than rare ones.\footnote{This latter similarity is
    reminiscent of \emph{weighted alternating least squares}
    \citep{wals}, which treats $f(x_{ij})$ as a confidence estimate
    that favors accurate estimation of certain parameters over
    uncertain ones.}

\end{itemize}

\citet{lessonslearned} note these algorithmic similarities. In their
controlled empirical comparison of several different embedding
approaches, results produced by SGNS and GloVe differ only
modestly.

There are subtle differences, however. The negative sampling regime of
SGNS ensures that the model does not place features near to one
another in the embedding space whose co-occurrence isn't observed in
the corpus.  This is distinctly different from GloVe, which trains
only on the \emph{observed} co-occurrence statistics. The GloVe model
incurs no penalty for placing features near to one another whose
co-occurrence has not been observed. As we shall see in Section 4,
this can result in poor estimates for uncommon features.

\section{Swivel}

Swivel is an attempt to have our cake and eat it, too.  Like GloVe, it
works from co-occurrence statistics rather than by sampling; like
SGNS, it makes use of the fact that many co-occurrences are
\emph{un}observed in the corpus.  Like both, Swivel performs a
weighted approximate matrix factorization of the PMI between
features. Furthermore, Swivel is designed to work well in a
distributed environment; e.g., \texttt{distbelief} \citep{distbelief}.

At a high level, Swivel begins with an $m \times n$ co-occurrence
matrix between $m$ row and $n$ column features. Each row feature and
each column feature is assigned a $d$-dimensional embedding
vector. The vectors are grouped into blocks, each of which defines a
submatrix ``shard''.  Training proceeds by selecting a shard (and
thus, its corresponding row block and column block), and performing a
matrix multiplication of the associated vectors to produce an estimate
of the PMI values for each co-occurrence. This is compared with the
observed PMI, with special handling for the case where no
co-occurrence was observed and the PMI is undefined. Stochastic
gradient descent is used to update the individual vectors and minimize
the difference.

As will be discussed in more detail below, splitting the matrix into
shards allows the problem to be distributed across many workers in a
way that allows for utilization of high-performance vectorized
hardware, amortizes the overhead of transferring model parameters, and
distributes parameter updates evenly across the feature embeddings.

\subsection{Construction}

To begin, an $m \times n$ co-occurrence matrix $\mathbf{X}$ is
constructed, where each cell $x_{ij}$ in the matrix contains the
observed co-occurrence count of row feature $i$ with column feature
$j$. The marginal counts of each row feature ($x_{i*} = \sum_j
x_{ij}$) and each column feature ($x_{*j} = \sum_i x_{ij}$) are
computed, as well as the overall sum of all the cells in the matrix,
$\lvert D \rvert = \sum_{i,j} x_{ij}$. As with other embedding
methods, Swivel is agnostic to both the domain from which the features
are drawn, and to the exact set of features that are
used. Furthermore, the ``feature vocabulary'' used for the rows need
not necessarily be the same as that which is used for the columns.

The rows are sorted in descending order of feature frequency, and are
then collected into $k$-element row blocks, where $k$ is chosen based
on computational efficiency considerations discussed below. This
results in $m/k$ row blocks whose elements are selected by choosing
rows that are congruent mod $m/k$. For example, if there are $2^{25}$
total rows in the co-occurrence matrix, for $k = 4096$, every $2^{25}
/ 4096 = 8,192^{th}$ row is selected to form a row block: the first
row block contains rows $(0, 8192, 16384, ...)$, the second row block
contains rows $(1, 8193, 16385, ...)$, and so on. Since rows were
originally frequency-sorted, this construction results in each row
block containing a mix of common and rare row features.

The process is repeated for the columns to yield $n/k$ column
blocks. As with the row blocks, each column block contains a mix of
common and rare column features.

For each (row block, column block) pair $(i, j)$, we construct a $k
\times k$ submatrix shard $\mathcal{X}_{ij}$ from the original
co-occurrence matrix by selecting the appropriate co-occurrence cells:

\[
\mathcal{X}_{ij} =
\left[ \begin{array}{cccc}
x_{i,j} & x_{i+\frac{m}{k},j} & \dots & x_{i+(k-1)\frac{m}{k},j} \\
x_{i,j+\frac{n}{k}} \\
x_{i,j+2\frac{n}{k}} \\
\vdots & & \ddots \\
x_{i,j+(k-1)\frac{n}{k}}
\end{array} \right]
\]

\begin{figure}[h]
  \includegraphics[width=\linewidth]{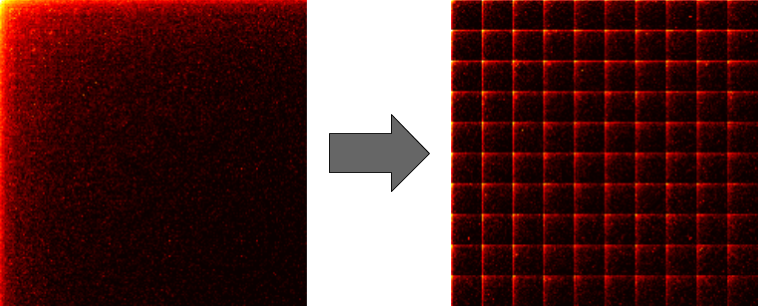}
  \caption{The matrix re-organization process creates shards with a
    mixture of common and rare features, which naturally leads to a
    mixture of large and small co-occurrence values (brighter and
    darker pixels, respectively).}
  \label{fig:shuf}
\end{figure}

This results in $mn/k^2$ shards in all. Typically, the vast majority
of these elements are zero. Figure \ref{fig:shuf} illustrates this
process: lighter pixels represent more frequent co-occurrences, which
naturally tend to occur for more frequent features.

\subsection{Training}

Prior to training, the two $d$-dimensional feature embeddings are
initialized with small, random values.\footnote{We specifically used
  values drawn from $\mathcal{N}(0, \sqrt{d})$: the choice was
  arbitrary and we did not investigate the effects of other
  initialization schemes.} $\mathbf{W} \in \mathbb{R}^{m \times d}$ is
the matrix of embeddings for the $m$ row features (e.g., words),
$\mathbf{\tilde{W}} \in \mathbb{R}^{n \times d}$ is the matrix of
embeddings for the $n$ column features (e.g., word contexts).

\begin{figure}[h]
  \includegraphics[width=\linewidth]{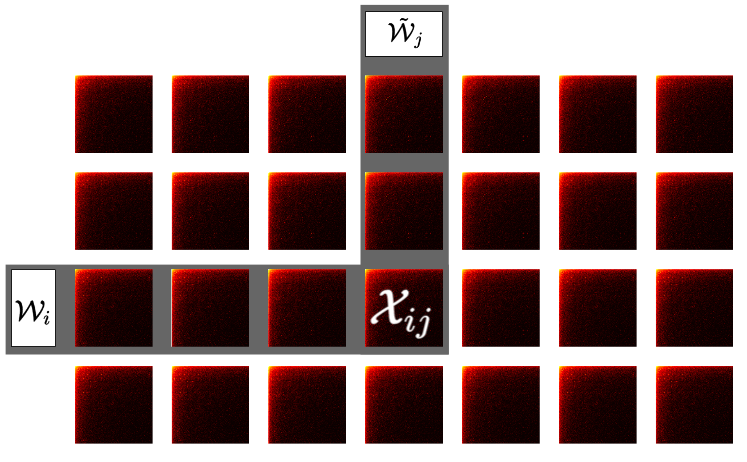}
  \caption{A shard $\mathcal{X}_{ij}$ is selected for training. The
    corresponding row vectors $\mathcal{W}_i$ and column vectors
    $\mathcal{\tilde{W}}_j$ are multiplied to produce estimates that
    are compared to the observed PMI derived from the count
    statistics.  Error is computed and back-propagated.}
  \label{fig:matmul}
\end{figure}

Training then proceeds iteratively as follows. A submatrix shard
$\mathcal{X}_{ij}$ is chosen at random, along with the $k$ row
embedding vectors $\mathcal{W}_i \in \mathbf{W}$ from row block $i$,
and the $k$ column vectors $\mathcal{\tilde{W}}_j \in
\mathbf{\tilde{W}}$ from column block $j$. The matrix product
$\mathcal{W}_i \mathcal{\tilde{W}}_j^\top$ is computed to produce
$k^2$ predicted PMI values, which are then compared to the observed
PMI values for shard $\mathcal{X}_{ij}$. The error between the
predicted and actual values is used to compute gradients: these are
accumulated for each row and column. Figure \ref{fig:matmul}
illustrates this process. The gradient descent is dampened using
Adagrad \citep{adagrad}, and the process repeats until the error no
longer decreases appreciably.

Although each shard is considered separately, it shares the embedding
parameters $\mathcal{W}_i$ for all other shards in the same row block,
and the embedding parameters $\mathcal{\tilde{W}}_j$ for all the other
shards in the same column block.  By storing the parameters in a
central \emph{parameter server} \citep{distbelief}, it is possible to
distribute training by processing several shards in parallel on
different worker machines.  An individual worker selects a shard,
retrieves the appropriate embedding parameters, performs the cost and
gradient computation, and then communicates the parameter updates back
to the parameter server. We do this in a lock-free fashion
\citep{hogwild} using Google's asynchronous stochastic gradient
descent infrastructure \texttt{distbelief} \citep{distbelief}.

Even on a very fast network, transferring the parameters between the
parameter server and a worker machine is expensive: for each $k \times
k$ block, we must retrieve (and then update) $k$ parameter values each
for $\mathcal{W}_i$ and $\mathcal{\tilde{W}}_j$.  Fortunately, this
cost is amortized over the $k^2$ individual estimates that are
computed by the matrix multiplication.  Choosing a reasonable value
for $k$ is therefore a balancing act between compute and network
throughput: the larger the value of $k$, the more we amortize the cost
of communicating with the parameter server. And up to a point,
vectorized matrix multiplication is essentially a constant-time
operation on a high-performance GPU. Clearly, this is all heavily
dependent on the underlying hardware fabric; we achieved good
performance in our environment with $k=4096$.

\subsection{Training Objective and Cost Function}

\begin{figure}[h]
  \includegraphics[width=\linewidth]{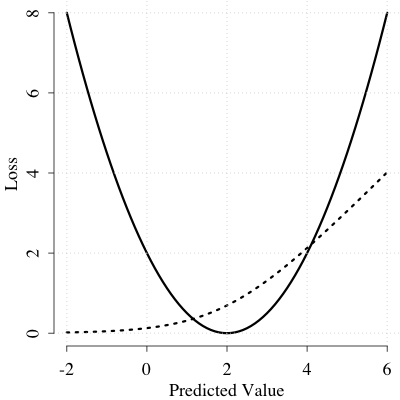}
  \caption{Loss as a function of predicted value $w_i
    \tilde{w}_j^\top$, evaluated for arbitrary objective value of
    $2.0$. The solid line shows $\mathcal{L}_1$ (squared error); the
  dotted line shows $\mathcal{L}_0$ (the ``soft hinge'').}
  \label{fig:loss}
\end{figure}

Swivel approximates the observed PMI of row feature $i$ and column
feature $j$ with $w_i^\top \tilde{w}_j$. It uses a piecewise loss
function that treats observed and unobserved co-occurrences
distinctly. Table \ref{tab:loss} summarizes the piecewise cost
function, and Figure \ref{fig:loss} shows the different loss function
variants with respect to $w_i^\top \tilde{w}_j$ for an arbitrary
objective value of 2.0.

\begin{table*}[t]
  \setlength{\tabcolsep}{8pt}
  \renewcommand{\arraystretch}{1}
  \centering
  \caption{Training objective and cost functions. The $\mathbf{pmi^*}$
    function refers to the ``smoothed'' PMI function described in the
    text, where the actual value of $0$ for $x_{ij}$ is replaced with $1$.}
  \resizebox{\textwidth}{!}{
  \begin{tabular}{c | c | p{8cm}}
    \hline
    \thead{Count}
    & \thead{Cost}
    & \thead{Intuition}
    
    \\
    \hline
    
    $x_{ij} > 0$
    & \gape{$\dfrac{1}{2}f(x_{ij})(w_i^\top \tilde{w}_j - \mathbf{pmi}(i;j))^2$}
    & Squared error: the model must accurately reconstruct observed PMI subject to our confidence in $x_{ij}$.
    \\
    \hline
    
    $x_{ij} = 0$
    & \gape{$\log \left[ 1 + \exp \left( w_i^\top \tilde{w}_j - \mathbf{pmi^*}(i;j) \right) \right]$}
    & ``Soft hinge:'' the model must not \emph{over}-estimate the PMI of common features whose co-occurrence is unobserved.
    \\
    \hline
  \end{tabular}}
  \label{tab:loss}
\end{table*}

\textbf{Observed co-occurrences.} For co-occurrences that have been
observed ($x_{ij} > 0$), we'd like $w_i^\top w_j$ to accurately
estimate $\mathbf{pmi}(i;j)$ subject to how confident we are in the
observed count $x_{ij}$. Swivel computes the weighted squared error
between the embedding dot product and the PMI of feature $i$ and
feature $j$:

\begin{align*}
  \mathcal{L}_1(i,j)
  &= \frac{1}{2} f(x_{ij}) \left( w_i^\top \tilde{w}_j - \mathbf{pmi}(i;j) \right)^2 \\
  &= \frac{1}{2} f(x_{ij}) (w_i^\top \tilde{w}_j - \log x_{ij} - \log \lvert D \rvert \\
  & \;\;\;\;\;\;\;\;\;\;\;\;\;\;\;\;\;\;\;\;\;\;\;\; + \log x_{i*} + \log x_{*j} )^2
\end{align*}

This encourages $w_i^\top \tilde{w}_j$ to correctly estimate the
observed PMI, as Figure \ref{fig:loss} illustrates. The loss is
modulated by a monotonically increasing confidence function
$f(x_{ij})$: the more frequently a co-occurrence is observed, the more
the model is required to accurately approximate
$\mathbf{pmi}(i;j)$. We experimented with several different variants
for $f(x_{ij})$, and discovered that a linear transformation of
$x_{ij}^{1/2}$ produced good results.

\textbf{Unobserved Co-occurrences.} Unfortunately, if feature $i$ and
feature $j$ are never observed together, $x_{ij} = 0$,
$\mathbf{pmi}(i; j) = -\infty$, and the squared error cannot be
computed.

What would we like the model to do in this case? Treating $x_{ij}$ as
a sample, we can ask: how \emph{significant} is it that its observed
value is zero? If the two features $i$ and $j$ are rare, their
co-occurrence could plausibly have gone unobserved due to the fact
that we simply haven't seen enough data. On the other hand, if
features $i$ and $j$ are common, this is less likely: it becomes
significant that a co-occurrence \emph{hasn't} been observed, so
perhaps we ought to consider that the features are truly
anti-correlated. In either case, we certainly don't want the model to
over-estimate the PMI between features, and so we can encourage the
model to respect an \emph{upper bound} on its PMI estimate $w_i^\top w_j$.

We address this by smoothing the PMI value as if a single
co-occurrence had been observed (i.e., computing PMI as if $x_{ij} =
1$), and using an asymmetric cost function that penalizes
over-estimation of the smoothed PMI. The following ``soft hinge'' cost
function (plotted as the dotted line in Figure \ref{fig:loss})
accomplishes this:

\begin{align*}
  \mathcal{L}_0(i,j) &= \log \; [ 1 + \exp(w_i^\top \tilde{w}_j - \mathbf{pmi^*}(i;j)) ] \\
  &= \log \; [ 1 + \exp( w_i^\top \tilde{w}_j - \log \lvert D \rvert \\
    & \;\;\;\;\;\;\;\;\;\;\; + \log x_{i*} + \log x_{*j}) ]
\end{align*}

Here, $\mathbf{pmi^*}$ refers to the smoothed PMI computation where
$x_{ij}$'s actual count of $0$ is replaced with $1$. This loss
penalizes the model for over-estimating the objective value; however,
it applies negligible penalty -- i.e., is non-committal -- if the
model under-estimates it.

Numerically, $\mathcal{L}_0$ behaves as follows. If features $i$ and
$j$ are common, the marginal terms $x_{i*}$ and $x_{*j}$ are large. In
order to minimize the loss, the model must produce a small -- or even
negative -- value for $w_i^\top \tilde{w}_j$, thus capturing the
anti-correlation between features $i$ and $j$. On the other hand, if
features $i$ and $j$ are rare, then the marginal terms are also small,
so the model is allowed much more latitude with respect to $w_i^\top
\tilde{w}_j$ before incurring serious penalty. In this way, the ``soft
hinge'' loss enforces an upper bound on the model's estimate for
$\mathbf{pmi}(i;j)$ that reflects the our confidence in the unobserved
co-occurrence.

\section{Experiments}

\begin{table*}[t]
\centering
\caption{Performance of SGNS, GloVe, and Swivel vectors across
  different tasks with respect to methods tested by
  \citet{lessonslearned}, with CBOW is included for reference. Word
  similarity tasks report Spearman's $\rho$ with human annotation;
  analogy tasks report accuracy.  In all cases, larger numbers
  indicate better performance.}  \resizebox{\textwidth}{!}{
\begin{tabular}{l | cccccc | cc }
  \hline
  Method
  & \thead{WordSim \\ Similarity}
  & \thead{WordSim \\ Relatedness}
  & \thead{\citeauthor{men} \\ MEN}
  & \thead{\citeauthor{mturk} \\ M. Turk}
  & \thead{\citeauthor{rarewords} \\ Rare Words}
  & \thead{\citeauthor{simlex999} \\ SimLex}
  & \thead{Google}
  & \thead{MSR} \\

  \hline

  SGNS  & \textrm{0.737} & \textrm{0.592} & \textrm{0.743} & \textrm{0.686} & \textrm{0.467} & \textrm{0.397} & \textrm{0.692} & \textrm{0.592} \\
  GloVe & \textrm{0.651} & \textrm{0.541} & \textrm{0.738} & \textrm{0.627} & \textrm{0.386} & \textrm{0.360} & \textrm{0.716} & \textrm{0.578} \\
  Swivel & \textbf{0.748} & \textbf{0.616} & \textbf{0.762} & \textbf{0.720} & \textbf{0.483} & \textbf{0.403} & \textbf{0.739} & \textbf{0.622} \\

  \hline

  CBOW & \textrm{0.700} & \textrm{0.527} & \textrm{0.708} & \textrm{0.664} & \textrm{0.439} & \textrm{0.358} & \textrm{0.667} & \textrm{0.570} \\

  \hline
\end{tabular}}
\label{tab:results}
\end{table*}

We performed several experiments to evaluate the embeddings produced
by Swivel.

\textbf{Corpora.} Following \citet{glove}, we produced 300-dimensional
embeddings from an August 2015 Wikipedia dump combined with the
Gigaword5 corpus. The corpus was tokenized, lowercased, and split into
sentences. Punctuation was discarded, but hyphenated words and
contractions were preserved. The resulting training data included 3.3
billion tokens across 89 million sentences. The most frequent 397,312
unigram tokens were used to produce the vocabulary, and the same
vocabulary is used for all experiments.

\textbf{Baselines.} In order to ensure a careful comparison, we
re-created embeddings using these corpora with the publicly-available
\texttt{word2vec}\footnote{\texttt{https://code.google.com/p/word2vec}}
and GloVe\footnote{\texttt{http://nlp.stanford.edu/projects/glove}}
programs as our baselines.

\texttt{word2vec} was configured to generate skipgram embeddings using
five negative samples. We set the window size so that it would include
ten tokens to the left of the focus and ten tokens to the right. We
ran it for five iterations over the input corpus. Since the least
frequent word in the corpus occurs 65 times, training samples the
rarest words at least 300 times each.  Since the same vocabulary is
used for both word and context features, we modified \texttt{word2vec}
to emit both word and context embeddings.  We experimented with adding
word vector $w_i$ with its corresponding context vector $\tilde{w}_i$
\citep{glove}; however, best performance was achieved using the word
vector $w_i$ alone, as was originally reported by \citet{word2vec}.

GloVe was similarly configured to use its ``symmetric'' co-occurrence
window that spans ten tokens to the left of the focus word and ten
tokens to the right. We ran GloVe for 100 training epochs using the
default parameter settings for initial learning rate ($\eta = 0.05$),
the weighting exponent ($\alpha = 0.75$), and the weighting function
cut-off ($x_{\mathbf{max}} = 100$). GloVe produces both word and
context vectors: unlike \texttt{word2vec}, the sum of the word vector
$w_i$ with its corresponding context vector $\tilde{w}_i$ produced
slightly better results than the word vector alone. (This was also
noted by \citet{glove}.)

Our results for these baselines vary slightly from those reported
elsewhere. We speculate that this may be due to differences in
corpora, preprocessing, and vocabulary selection, and simply note
that this evaluation should at least be internally consistent.

\textbf{Swivel Training.} The unigram vocabulary was used for both row
and column features. Co-occurrence was computed by examining ten words
to the left and ten words to the right of the focus word. As with
GloVe, co-occurrence counts were accumulated using a harmonically
scaled window: for example, a token that is three tokens away from the
focus was counted as $\frac{1}{3}$ of an occurrence.\footnote{We
  experimented with both linear and uniform scaling windows, and
  neither performed as well.}  So it turns out that GloVe and Swivel
were trained from identical co-occurrence
statistics.\footnote{Following \citeauthor{matrixfactorization}'s
  \citeyear{matrixfactorization} suggestion that SGNS factors a
  shifted PMI matrix, we experimented with shifting the objective PMI
  value by a small amount. Specifically, \citet{matrixfactorization}
  suggest that the SGNS PMI objective is shifted by $\log k$, where
  $k$ is the number of negative samples drawn from the unigram
  distribution. Since we'd configured \texttt{word2vec} with $k = 5$,
  we experimented with shifting the PMI objective by $\log 5$ ($\sim
  1.61$).  This did not yield significantly different results than
  just using the original PMI objective.}

We trained the model for a million ``steps'', where each step trains
an individual submatrix shard.  Given a vocabulary size of roughly
400,000 words and $k=4096$, there are approximately 100 row blocks and
100 column blocks, yielding 10,000 shards overall.  Therefore each
shard was sampled about 100 times.

We experimented with several different weighting functions to modulate
the squared error based on cell frequency of the form $f(x_{ij}) = b_0
+ b x_{ij}^\alpha$ and found that $\alpha = \frac{1}{2}$, $b_0 = 0.1$,
and $b = \frac{1}{4}$ yielded good results.

Finally, once the embeddings were produced, we discovered that adding
the word vector $w_i$ to its corresponding context vector
$\tilde{w}_i$ produced better results than the word vector $w_i$
alone, just as it did with GloVe.

\textbf{Evaluation.} We evaluated the embeddings using the same
datasets that were used by \citet{lessonslearned}. For word
similarity, we used WordSim353 \citep{wordsim353} partitioned into
WordSim Similarity and WordSim Relatedness
\citep{wordsimsim,wordsimrel}; \citeauthor{men}'s \citeyear{men} MEN
dataset; \citeauthor{mturk}'s \citeyear{mturk} Mechanical Turk,
\citeauthor{rarewords}'s \citeyear{rarewords} Rare Words; and
\citeauthor{simlex999}'s \citeyear{simlex999} SimLex-999
dataset. These datasets contain word pairs with human-assigned
similarity scores: the word vectors are evaluated by ranking the pairs
according to their cosine similarities and measuring the correlation
with the human ratings using Spearman's $\rho$. Out-of-vocabulary
words are ignored.

The analogy tasks present queries of the form ``A is to B as C is to
X'': the system must predict X from the entire vocabulary. As with
\citet{lessonslearned}, we evaluated Swivel using the MSR and Google
datasets \citep{msreval,word2vec}. The former contains syntactic
analogies (e.g., ``\emph{good} is to \emph{best} as \emph{smart} is to
\emph{smartest}''). The latter contains a mix of syntactic and
semantic analogies (e.g., ``\emph{Paris} is to \emph{France} as
\emph{Tokyo} is to \emph{Japan}''). The evaluation metric is the number of
queries for which the embedding that maximizes the cosine similarity
is the correct answer. As with \cite{word2vec}, any query terms are
discarded from the result set and out-of-vocabulary words are scored
as losses.

\textbf{Results.} The results are summarized in Table
\ref{tab:results}.  Embeddings produced by \texttt{word2vec}'s CBOW
are also included for reference.  As can be seen, Swivel outperforms
GloVe, SGNS, and CBOW on both the word similarity and analogy
tasks. We also note that, except for the Google analogy task, SGNS
outperforms GloVe.

Our hypothesis is that this occurs because both SGNS and Swivel take
unobserved co-occurrences into account, but GloVe does not.  Swivel
incorporates information about unobserved co-occurrences directly,
including them in among the predictions and applying the ``soft
hinge'' loss to avoid over-estimating the feature pair's PMI. SGNS
indirectly models unobserved co-occurrences through negative sampling.
GloVe, on the other hand, only trains on positive co-occurrence data.

We hypothesize that by not taking the unobserved co-occurrences into
account, GloVe is under-constrained: there is no penalty for placing
unobserved but unrelated embeddings near to one
another. Quantitatively, the fact that both SGNS and Swivel
out-perform GloVe by a large margin on \citeauthor{rarewords}'s
\citeyear{rarewords} Rare Words evaluation seems to support this
hypothesis. Inspection of some very rare words (Table \ref{tab:fold})
shows that, indeed, SGNS and Swivel have produced reasonable
neighbors, but GloVe has not.

\begin{table*}[t]
  \setlength{\tabcolsep}{8pt}
  \renewcommand{\arraystretch}{1}
  \centering
  \caption{Nearest neighbors for some very rare words.}
  \resizebox{\textwidth}{!}{
  \begin{tabular}{l | c | >{\raggedright}p{6cm} | >{\raggedright}p{6cm} | >{\raggedright}p{6cm} c}
    \hline
    \thead{Query}
    & \thead{Vocabulary Rank}
    & \thead{SGNS}
    & \thead{GloVe}
    & \thead{Swivel}
    &
    \\
    
    \hline\hline

    bootblack
    & 393,709
    & shoeshiner, newsboy, shoeshine, stage-struck, bartender, bellhop, waiter, housepainter, tinsmith
    & redbull, 240, align=middle, 18, 119, dannit, concurrence/dissent, 320px, dannitdannit
    & newsboy, shoeshine, stevedore, bellboy, headwaiter, stowaway, tibbs, mister, tramp
    &
    \\

    \hline
    
    chigger
    & 373,844
    & chiggers, webworm, hairballs, noctuid, sweetbread, psyllids, rostratus, narrowleaf, pigweed
    & dannit, dannitdannit, upupidae, bungarus, applause., .774, amolops, maxillaria, paralympic.org 
    & mite, chiggers, mites, batatas, infestation, jigger, infested, mumbo, frog's
    &
    \\

    \hline
    
    decretal
    & 374,123
    & decretals, ordinatio, sacerdotalis, constitutiones, theodosianus, canonum, papae, romanae, episcoporum
    & regesta, agatho, afl.com.au, dannitdannit, dannit, emptores, beatifications, 18, 545
    & decretals, decretum, apostolicae, sententiae, canonum, unigenitus, collectio, fidei, patristic
    &
    \\

    \hline

    tuxedoes
    & 396,973
    & tuxedos, ballgowns, tuxes, well-cut, cable-knit, open-collared, organdy, high-collared, flouncy
    & hairnets, dhotis, speedos, loincloths, zekrom, shakos, mortarboards, caftans, nightwear
    & ballgowns, tuxedos, tuxes, cummerbunds, daywear, bridesmaids', gowns, strapless, flouncy
    &
    \\
    \hline\hline

  \end{tabular}}
  \label{tab:fold}
\end{table*}

\begin{figure}[h]
  \includegraphics[width=\linewidth]{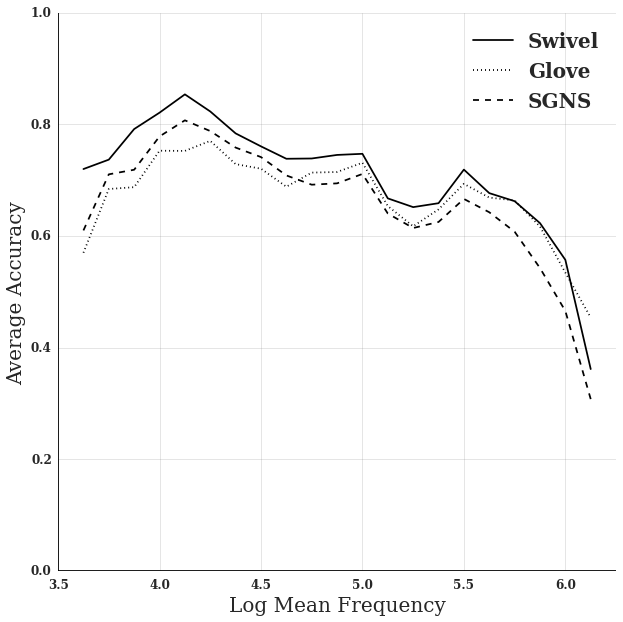}
  \caption{Analogy accuracy as a function of the log mean frequency of the four words.}
  \label{fig:freq}
\end{figure}

To be fair, GloVe was explicitly designed to capture the relative
geometry in the embedding space: the intent was to optimize for
performance on analogies rather than on word similarity. Nevertheless,
we see that word frequency has a marked effect on analogy performance,
as well. Figure \ref{fig:freq} plots analogy task accuracy against the
base-10 log of the mean frequency of the four words involved.

To produce the plot, we considered both the MSR and Google
analogies. For each analogy, we computed the mean frequency of the
four words involved, and then bucketed it with other analogies that
have similar mean frequencies. Each bucket contains at least 100
analogies.

Notably, Swivel performs better than SGNS at all word frequencies, and
better than GloVe on all but the most frequent words. GloVe
under-performs SGNS on rare words, but begins to out-perform SGNS as
the word frequency increases. We hypothesize that GloVe is fitting the
common words at the expense of rare ones.

It is also interesting to note that all algorithms tend to perform
poorly on the most frequent words. This is probably because very
frequent words a) tend to appear in many contexts, making it difficult
to determine an accurate point representation, and b) they tend to be
polysemous, appear as both verbs and nouns, and have subtle gradations
in meaning (e.g., \emph{man} and \emph{time}).

\section{Discussion}

Swivel grew out of a need to build embeddings over larger feature sets
and more training data. We wanted an algorithm that could both handle
a large amount of data, and produced good estimates for both common
and rare features.

\textbf{Statistics vs. Sampling.} Like GloVe, Swivel trains from
\emph{co-occurrence statistics}: once the co-occurrence matrix is
constructed, training Swivel requires computational resources in
proportion to the matrix size.  This allows Swivel to handle much more
data than can be practically processed with a sampling method like
SGNS, which requires training time in proportion to the size of the
corpus.

\textbf{Unobserved Co-occurrences.} Our experiments indicate that
GloVe pays a performance cost for only training on \emph{observed}
co-occurrences. In particular, the model may produce unstable
estimates for rare features since there is no penalty for placing
features near one another whose co-occurrence isn't
observed.

Nevertheless, computing values for \emph{every} pair of features
potentially entails significantly more computation than is required by
GloVe, whose training complexity is proportional to the number of
\emph{non-zero} entries in the co-occurrence matrix. Swivel mitigates
this in two ways.

First, it makes use of vectorized hardware to perform matrix
multiplication of thousands of embedding vectors at once. Performing
about a dozen $4096 \times 4096$ matrix multiplications per GPU
compute unit per second is typical: we have observed that a single GPU
can estimate about 200 million cell values per second for
1024-dimensional embedding vectors.

Second, the blocked matrix shards can be separately processed by
several worker machines to allow for coarse-grained parallelism. The
block structure amortizes the overhead of transferring embedding
parameters to and from the parameter server across millions of
individual estimates. We found that Swivel did, in fact, parallelize
easily in our environment, and have been able to run experiments that
use hundreds of concurrent worker machines.

\textbf{Piecewise Loss.} It seems fruitful to consider the
co-occurrence matrix as itself containing estimates rather than point
values. A corpus is really just a sample of language, and so a
co-occurrence matrix \emph{derived} from a corpus itself contains
samples whose values are uncertain.

We used a weighted piecewise loss function to capture this
uncertainty.  If a co-occurrence was observed, we can produce a PMI
estimate, and we require the model to fit it more or less accurately
based on the observed co-occurrence frequency.  If a co-occurrence was
not observed, we simply require that the model avoid
\emph{over}-estimating a smoothed PMI value. While this works well, it
does seem \emph{ad hoc}: we hope that future investigation can yield a
more principled approach.

\section{Conclusion}

Swivel produces low-dimensional feature embeddings from a
co-occurrence matrix. It optimizes an objective that is very similar
to that of SGNS and GloVe: the dot product of a word embedding with a
context embedding ought to approximate the observed PMI of the two
words in the corpus.

Unlike SGNS, Swivel's computational requirements depend on the size of
the co-occurrence matrix, rather than the size of the corpus. This
means that it can be applied to much larger corpora.

Unlike GloVe, Swivel explicitly considers \emph{all} the co-occurrence
information -- including unobserved co-occurrences -- to produce
embeddings.  In the case of unobserved co-occurrences, a ``soft
hinge'' loss prevents the model from over-estimating PMI. This leads
to demonstrably better embeddings for rare features without
sacrificing quality for common ones.

Swivel capitalizes on vectorized hardware, and uses block structure to
amortize parameter transfer cost and avoid contention. This results in
the ability to handle very large co-occurrence matrices in a scalable
way that is easy to parallelize.

We would like to thank Andrew McCallum, Samy Bengio, and Julian
Richardson for their thoughtful comments on this work.
\bibliography{swivel}
\bibliographystyle{plainnat}
\end{document}